\documentclass{Interspeech}

\usepackage[T1]{fontenc}
\usepackage{glossaries}
\usepackage{multirow}
\usepackage{makecell}
\usepackage{multicol}
\usepackage{hyperref}
\usepackage{cleveref}
\usepackage[normalem]{ulem}
\usepackage{pgfplots}
\usepackage{graphicx}
\makeglossaries
\glsdisablehyper

\newacronym{ASR}{ASR}{automatic speech recognition}
\newacronym{WER}{WER}{word error rate}
\newacronym{CTC}{CTC}{connectionist temporal classification}
\newacronym{KL}{KLD}{Kullback-Leibler divergence}
\newacronym{KD}{KD}{knowledge distillation}
\newacronym{LM}{LM}{language model}

\newacronym[first=blank elimination]{BE}{blank elimination}{blank elimination}

\newacronym{TED}{TEDv2}{TED-LIUMv2}
\newacronym{LBS}{LBS}{LibriSpeech}

\interspeechcameraready 

\title{Analyzing the Importance of Blank for CTC-Based\\ Knowledge Distillation}

\author[affiliation={1,2}]{Benedikt}{Hilmes}
\author[affiliation={1,2}]{Nick}{Rossenbach}
\author[affiliation={1,2}]{Ralf}{Schlüter}

\affiliation{RWTH Aachen University}{Aachen}{Germany}
\affiliation{Apptek GmbH}{Aachen}{Germany}
\email{<lastname>@ml.rwth-aachen.de}
\keywords{speech recognition, knowledge distillation, blank distribution, foundation models}

\usepackage{comment}

\begin{document}

\maketitle

\begin{abstract}
With the rise of large pre-trained foundation models for automatic speech recognition new challenges appear.
While the performance of these models is good, runtime and cost of inference increases.
One approach to make use of their strength while retaining efficiency is to distill their knowledge to smaller models during training.
In this work, we explore different CTC-based distillation variants, focusing on blank token handling.
We show that common approaches like blank elimination do not always work off the shelf.
We explore new blank selection patterns as a potential sweet spot between standard knowledge distillation and blank elimination mechanisms.
Through the introduction of a symmetric selection method, we are able to remove the CTC loss during knowledge distillation with minimal to no performance degradation. With this, we make the training independent from target labels, potentially allowing for distillation on untranscribed audio data.

\end{abstract}

\section{Introduction}
Recently, foundation models emerged as the new state-of-the-art in \gls{ASR}.
These models are trained on large amounts of data either in a supervised or unsupervised fashion.
Examples for these large pre-trained models are Whisper \cite{Radford2022RobustSR}, Hubert \cite{Hsu2021HuBERTSS} or Wav2Vec \cite{wav2vec2}.
While the training methods differ, they perform well on a number of downstream tasks.
When deploying models to mobile or edge devices, resources are limited and inference speed as well as memory footprint are a major limitation of foundation models.
Still, it is desirable to use the capabilities provided by these models, as their performance is often superior to a smaller and directly trained model.
One way to do that is through \gls{KD}, where a larger teacher model serves to improve a smaller student model through training the model both on the given data and the output distribution of the teacher model.

There has been quite some work done in order to tackle the challenges of distilling the knowledge of \gls{CTC}-based \gls{ASR} models.
\gls{KD} was initially proposed in \cite{Hinton2015DistillingTK} as a general method to transfer information from one neural network to another.
One of the problems of applying \gls{KD} to \gls{CTC} is the amount of blanks in the predictions of the models.
To deal with this, \cite{Takashima2018AnIO} and \cite{huang18d_interspeech} propose sequence level \gls{KD}, as an alternative to frame level \gls{KD}.
In \cite{Kurata2019GuidingCP}, the student is trained in a way that the \gls{CTC} spike timings align with those of the teacher.
This alignment problem is taken up in \cite{8639629}, which proposed to counteract the mismatch in alignment between teacher and student distributions through distilling over frame areas.
\cite{Tian2022KnowledgeDF}~introduces a two-stage training, where first the features of the student are trained to match those of the teacher.
In a second step, they calculate the distillation loss by combining a \gls{KL} loss on the non-blank positions of the teacher with a \gls{KL} loss tying the student and teacher features together.
In \cite{TIAN2024103071}, blank and non-blank distributions are factorized separately and combined in a three stage process: 1. Binary distinction between blank and non-blank; 2. Balanced distillation on blank and non-blank; 3. Learning a posterior matrix over a factorized \gls{KD} loss.

For architectures like RNN-T, distilling the model on the teachers outputs only is more common \cite{Panchapagesan2020EfficientKD, zeineldeen2023:rnnt_fullsum_kd}.
Most \gls{CTC}-based approaches combine the \gls{KD} objective with the \gls{CTC} objective \cite{Tian2022KnowledgeDF, TIAN2024103071}.
This is why as a baseline for this work we use a setup that includes both the \gls{CTC} and \gls{KD} objective, and confirm that for base \gls{KD} the \gls{CTC} loss is helpful.
A potentially undesired side-effect of this combination is the dependence on labeled data for the \gls{CTC} loss.
While one solution would be to have the teacher generate pseudo labels as targets, we refrain from this approach, as this has been shown not to be helpful when doing distillation \cite{zeineldeen2023:rnnt_fullsum_kd}.
Usually, work on \gls{CTC}-based \gls{KD} starts with a medium sized teacher to distill the information to a small student, resulting in maximum size reduction for the final model \cite{Takashima2018AnIO, huang18d_interspeech, Tian2022KnowledgeDF, TIAN2024103071}.
In this work we focus on keeping the student medium sized and instead try to deploy a larger foundation model as teacher, pre-trained on unlabeled data, in order to provide maximum information to the student.
We do this with the goal of having an efficiently deployable model with performance enhanced by a large foundation model teacher.

\vspace{-0.5em}
\subsection{Contribution}
\vspace{-0.5em}
Our contributions are as follows: We start from investigating the behavior of blank elimination \cite{Tian2022KnowledgeDF} for \gls{CTC}-based \gls{KD}.
We then introduce different extensions with the goal of using the information contained in the blank predictions of our teacher.
We analyze these approaches in training settings with differently scaled distillation losses.
One of these approaches, symmetric blank selection, includes only the neighboring positions of non-blank positions.
We found that with this particular subset of blanks we can remove the \gls{CTC} objective from the distillation process without losing performance.
This is in contrast to the standard blank elimination and other newly introduced approaches, where the performance can not be maintained without the CTC component.
With a distillation process using symmetric blank selection and dropping the \gls{CTC} loss, it becomes possible to include unsupervised data in the distillation process.
We run experiments on two English corpora, with baseline models close to state-of-the-art by using HuBERT as a large foundation model as teacher.
Our work and software \cite{DBLP:conf/emnlp/PeterBN18, DBLP:conf/icassp/DoetschZVKSN17} is public.\footnote{https://github.com/rwth-i6/returnn-experiments/tree/master/2025-ctc-blank-kd}

\section{Connectionist Temporal Classification}
\label{sec:ctc}
\glsreset{CTC}
In this work we consider \gls{ASR} models using \gls{CTC} \cite{10.1145/1143844.1143891}.
CTC is used to align an acoustic signal $x_1^T$ and a label sequence $a_1^S$
through the introduction of a blank label, which extends $a_1^S$ to length $T$.
An alignment $y_1^T$ is a valid alignment for $x_1^T$ and $a_1^S$ iff $B(y_1^T) = a_1^S$, where $B$ is a function that removes all blanks and consecutive label repetitions from the sequence.
A model can now be trained with the \gls{CTC} criterion by marginalizing over all possible alignments and minimizing the following objective:
\begin{equation}
\hspace{-0.35em}\mathcal{L}_{\text{CTC}} = -\log p(a_1^S | x_1^T) = -\log \sum_{y_1^T : B(y_1^T) = a_1^S} p(y_1^T | x_1^T) 
\end{equation}
It is a well known observation that \gls{CTC} produces alignments which are dominated by the blank label in what is called a ``peaky behavior'' \cite{zeyer2021:peakyctc, Liu2018ConnectionistTC, LI2020107392, Huang2024LessPA}.
This means that for the frame-wise predictions of a model trained with \gls{CTC} criterion, a majority of the probability mass lies in the position of the blank symbol.
While this is not hurtful during recognition, when distilling knowledge from a teacher to a student, some problems may occur, as will be discussed in the following section.

\vspace{-0.75em}
\glsreset{KD}
\section{Knowledge Distillation}
\Gls{KD} describes the process of transferring the information embedded in a so called teacher model into a (usually smaller) student model \cite{Hinton2015DistillingTK}. 
There are multiple variations of \gls{KD}, which range from pseudo-labeling unsupervised data via the teacher, to training the student to have a similar output distribution as the teacher.
Depending on the architecture of the student and teacher model different approaches have shown to be the most effective.

A major distinction is if the model output simultaneously serves as label-to-acoustic alignment, as is the case for CTC and RNN-T-based models.
For such models, minimizing the frame-level cross entropy error between the posterior distributions of the teacher and the student model is difficult \cite{Takashima2018AnIO}.

Instead, in this work we focus on distilling the knowledge of the teacher to the student by minimizing the \gls{KL} of the posterior distributions. \cite{Hinton2015DistillingTK,Tian2022KnowledgeDF,TIAN2024103071}.

The criterion for a given sequence can be formalized as:
\begin{equation}
	\label{eq:KL}
	\mathcal{L}_{\text{KD}} = \sum_{t=1}^{T} \sum_{c=1}^C p_{\text{tea}}(c | x_1^T, t) \cdot \log \frac{p_{\text{tea}}(c | x_1^T, t)}{p_{\text{stu}}(c | x_1^T, t)}
\end{equation}
with given acoustic signal $x_1^T$ of length $T$, $C$ classes including blank, teacher distribution $p_{\text{tea}}$ and student distribution $p_{\text{stu}}$.

\vspace{-0.75em}
\subsection{Blank Elimination}
\vspace{-0.25em}
The calculation of \Cref{eq:KL} includes both blank and non-blank positions of the teacher distribution. 
As stated in \Cref{sec:ctc}, \gls{CTC} tends to produce peaky alignments where a majority of the positions are blank.
Consequently, the influence of blank positions seen during training proportionally surpasses that of non-blank positions.
When viewed on a per-class comparison, this effect increases by another magnitude.
This means that making use of the \gls{KD} criterion without further modifications hinders the convergence of the distilled model and hurts performance, which was already observed in previous works \cite{Takashima2018AnIO, Takashima2019InvestigationOS, Senior2015AcousticMW}.
To deal with this problem \cite{Tian2022KnowledgeDF} proposes to calculate the \gls{KL} only on the non-blank positions of the sequence which we are going to call \textit{blank elimination} in the following.
These positions are calculated in a greedy fashion by looking at the posterior distribution of the teacher and selecting the positions where a non-blank token has the highest probability. 
The modified \gls{KD} criterion can be described as:
\vspace{-1.5em}

\begin{multline}
	\mathcal{L}_{\text{KD-BE}} = \sum_{t=1}^{T} \sum_{c=1}^C p_{\text{tea}}(c | x_1^T, t) \cdot \log (\frac{p_{\text{tea}}(c | x_1^T, t)}{p_{\text{stu}}(c | x_1^T, t)}) \\
	\cdot (1-\delta(\text{<b>}, \arg\max_{c'}\{p_{\text{tea}}(c' | x_1^T, t)\}))
\end{multline}
where $\delta$ describes the Kronecker delta and $\text{<b>}$ is the index of the blank token.
\vspace{-0.5em}
\subsection{Blank Selection Mechanisms}
\vspace{-0.25em}
\label{sec:blank_sel}
Learning a proper blank distribution is a crucial part of training and decoding a model with the \gls{CTC} criterion. 
Since the gradient of the \gls{KD} objective for the case of blank elimination is only computed on non-blank positions, an additional \gls{CTC} loss is used in prior work to update positions where the teacher does predict blank. 
This means that without further modification training requires supervised data and a larger portion of the teacher outputs is not considered for \gls{KD}.
As shown in \cite{Tian2022KnowledgeDF}, distilling knowledge over all blanks is hurtful for the convergence of the model.
Nevertheless, completely removing them from the distillation means that the teacher's distributions at blank positions, and consequently parts of the alignment, are not propagated to the student. 
Thus, in the following we explain novel approaches to consider only specific blank positions.
\vspace{-0.25em}
\subsubsection{Non-blank dependent selection}
\vspace{-0.25em}
For our first approach, we propose to extend the blank elimination symmetrically around the non-blank positions of the teacher output. 
This means that for every frame where the argmax of the teacher is not the blank symbol, the surrounding $n$ frames will also be considered for the calculation of the \gls{KL}.
This way, the blank information at the boundary is distilled, while at positions with larger amounts of sequential blank positions the predictions are left out.

As we intentionally only select positions around non-blank predictions of the teacher, we do not check if the selected position would already be included due to being non-blank.
While in theory the amount of blanks selected is upper bounded by $2n$, the amount of actually selected blank positions is less, which we are going to analyze in \Cref{sec:keepsome}.

The amount of blanks included into the distillation is a tunable parameter, which can differ for different corpora and models.
For our experiments we test a value of $n$ ranging from $1$ to $5$.
Since we hypothesize that within-sequence blanks might be more important than blank predictions outside of the sequence, we add a chain of experiments where we only trim the blanks before the first and after the last non blank prediction of the teacher from the \gls{KL} calculation.
\vspace{-0.25em}
\subsubsection{Probability-based selection}
\vspace{-0.25em}
Instead of relying on our non-blank positions to select which blank positions we include during training, we do a selection based on the probability distributions of the blank positions.
For this, we consider a threshold $\alpha$ and include all positions into our \gls{KD} process for which the probability of blank is below $\alpha$.
This way, we aim to select positions where the teacher is less sure about its blank prediction.
Thus, a potential transition into non-blank occurs, which might be an important information for the student to learn.
We try the values $0.95, 0.9$ and $0.8$ for $\alpha$.

As additional control experiment, we include a random sampling approach in our experiments, where the blank positions seen \mbox{during} \gls{KD} are selected randomly. 
Here we use a scale $\beta$ which decides how many positions proportionally to the non-blank positions are selected, where $\beta = 1$ means that we randomly select as many positions as there are non-blank positions in the teacher output.
For $\beta$ we test $0.5, 1.0$ and $2.0$.
\vspace{-0.75em}
\section{Experimental Setup}
\vspace{-0.5em}
\subsection{Data}
\vspace{-0.25em}
In this work we make use of the two English datasets \gls{TED} \cite{Rousseau2014EnhancingTT} and \gls{LBS} \cite{7178964}.
\gls{TED} consists of 207 hours of TED talks, while \gls{LBS} offers 960 hours of audio book recordings for training.
For LibriSpeech we report results on \textit{dev-other} and \textit{test-other}.
We split the data into sub-epochs of 5 for \gls{TED} and 10 for \gls{LBS}.
We use phonemes as targets, with the phoneme set consisting of ARPABET phoneme symbols without stress marker.
In order to predict phoneme sequences missing for words in the lexicon during training, we use Sequitur \cite{Bisani-2008-Joint-sequencemodel}.
For recognition, we use a 4-gram \gls{LM} trained on the corresponding text-only data provided with each of the corpora.
\vspace{-0.5em}
\subsection{Teacher}
\vspace{-0.25em}
As a teacher for this work, we use the publicly available \textit{large} variant of the HuBERT model\footnote{https://huggingface.co/facebook/hubert-large-ll60k} (317M parameters) \cite{Hsu2021HuBERTSS}, only trained on Libri-Light \cite{Kahn2019LibriLightAB}.
In order to adjust the model to the task, we add a linear layer at the end and fine-tune the whole model to each corpus for 10 full epochs.
Using the publicly available \gls{LBS} fine-tuned version of the HuBERT model did not yield a performance difference, so we stick to the unsupervised model.
We evaluate each full epoch and take the checkpoint with the best dev \gls{WER} as teacher checkpoint.
We down-sample the outputs by another factor of two in order to match the time resolution of the student.
We found that this does not result in a worse \gls{WER} for the teacher.
For knowledge distillation, we disable dropout and other data augmentation techniques, as this improves the \gls{KD} \cite{10096418}.
\vspace{-0.5em}
\subsection{Training}
\vspace{-0.25em}
Our student model is a 12-layer Conformer \cite{conformer} with relative positional encodings \cite{Shaw2018SelfAttentionWR}. 
The hidden dimension of our student is 384, going up to 1536 for the feed-forward module. 
In total the model consists of around 42M parameters.
We use 80-dimensional log-mel features with a frame shift of 10ms as input to our network, augmented by SpecAugment \cite{Park2019SpecAugmentAS}.
Our frontend consisting of a stack of convolutions downsamples the features by a factor of 4, so that each output label corresponds to 40ms.
We use a dropout \cite{JMLR:v15:srivastava14a} probability of 0.2 for \gls{TED} and 0.1 for \gls{LBS}. 
We apply the AdamW \cite{loshchilov2018decoupled} optimizer with a weight decay scale of 0.01. We use a linear one cycle learning rate scheduling starting from 7e-6 peaking at 5e-4 after 235 sub-epochs, going down to 5e-5 and finishing with a cooldown of 30 sub-epochs down to 1e-7. 
We train our models for 500 sub-epochs, so 50 full epochs for \gls{TED} and 100 epochs \gls{LBS}.
For \gls{TED} we use a batch size of 180 seconds of audio, while for \gls{LBS} we use 300 seconds.
We train our baseline with \gls{CTC} loss and use Flashlight \cite{kahn2022flashlight} for decoding, using the last checkpoint for recognition.
For training our models with knowledge distillation, the loss consists of a \gls{KL} loss with the teacher's posterior distribution as target, and an optional (for $\lambda < 1$) interpolation with a \gls{CTC} loss with the phoneme labels as targets:
\begin{equation}
	\mathcal{L} = \lambda \mathcal{L}_{KD} + (1 - \lambda) \mathcal{L}_{CTC}
\end{equation}
In initial experiments, we tried a number of different scales between 0 and 1. We found that 0.25, 0.9 and 1.0 give the best impression on the behavior of different \gls{KD} conditions, exhibiting the best results.
This is why we put our focus on these three settings: 1. Emphasis on \gls{CTC} loss (0.25); 2. Emphasis on \gls{KD} loss (0.9); 3. No \gls{CTC} loss (1.0).
All experiments can be run on a single consumer GPU with 24gb VRAM (e.g. Nvidia RTX 3090), resulting in a low barrier for reproduction.
\vspace{-0.5em}
\section{Experiments}
\subsection{Baselines}
\begin{table}
	\caption{Baseline performance of \gls{TED} and \gls{LBS} with 4-gram \gls{LM}. For LibriSpeech we evaluate on the \textit{other} portions of the evaluation data. A distillation scale of 1.0 means no CTC loss is used.}
	\vspace{-4ex}
	\label{tab:baselines}
	\begin{center}
			\begin{tabular}{|c|c||c|c||c|c|}
				\hline
				\multirow{3}{*}{Model}  & \multirow{3}{*}{\makecell{Distillation \\ Scale}} & \multicolumn{4}{c|}{WER [\%]}\\
				\cline{3-6}
				& & \multicolumn{2}{c||}{\gls{TED}} & \multicolumn{2}{c|}{\gls{LBS}} \\
				\cline{3-6}
				& & dev& test& dev & test \\
				\hline
				Teacher & - & 4.8 & 5.4 & 4.0 & 4.0\\
				\hline
				Baseline & - & 6.4 & 7.0 & 6.0 & 6.6\\
				\hline
				\hline
				\multirow{3}{*}{KD} & 0.25 & 6.6 & 7.1 & \textbf{5.6} & 6.1\\
				\cline{2-6}
				& 0.9 & 6.0 & 6.7 & 5.8 & \textbf{6.0}\\
				\cline{2-6}
				& 1.0 & 6.3 & 6.6 & 5.9 & 6.4\\
				\hline
				\multirow{3}{*}{\makecell{Blank \\ Elim.}} & 0.25 & 6.1 & 6.9 & 5.7 & 6.1\\
				\cline{2-6}
				& 0.9 & \textbf{5.8} & \textbf{6.5} & 6.1 & 6.5\\
				\cline{2-6}
				& 1.0 & 6.4 & 6.8 & 6.9 & 7.3\\
				\hline
			\end{tabular}
	\end{center}
	\vspace{-5ex}
\end{table}
The baseline performance of our models can be seen in \Cref{tab:baselines}. 
Comparing the teacher with our baseline trained without any \gls{KD}, we can see that the teacher outperforms the baseline by around 30\% \gls{WER} relative across multiple corpora and test sets. This is to be expected, as the teacher not only was trained on a lot more data, but also contains around 8 times the number of parameters compared to the baseline.

When adding \gls{KD} to train the student first differences become visible.
In general, both models benefit from the additional information given by the teacher, but different distillation scales seem to be optimal.
For \gls{TED} a higher scale (0.9 or 1.0) produces better results than the lower scale (0.25) which is only on par with the baseline.
For \gls{LBS} a lower scale seems to perform better, where the best result of 5.6\% on dev-other is achieved with a scale of 0.25. 
Nevertheless, the two \gls{LBS} models with higher scales also outperform the baseline.

When removing blank positions from the \gls{KD} objective, a major difference is visible. 
Blank elimination helps in the case of \gls{TED} where the \gls{WER} goes down to 5.8\% on the dev set, but for the models trained on \gls{LBS} performance degrades.
Especially for the case where the distillation scale is 1.0 this difference is strongly visible for \gls{LBS} dev-other, going up from 5.9\% to 6.9\%.
Since a distillation scale of 1.0 effectively disables the \gls{CTC} objective, this hints that the \gls{CTC} objective for the case of blank elimination is able to counteract losing teacher information about the blank positions.
Looking at \gls{TED}, this difference does not seem to be important as much, meaning that the blank positions for \gls{TED} are not as important as for \gls{LBS}.
We also experimented with prior correction, where the blank prior is removed from the teacher distribution, but results were consistently worse.
\vspace{-1.5ex}
\subsection{Blank Aware Approaches}
\label{sec:keepsome}
\begin{figure}
	\caption{\gls{WER} [\%] of different symmetric selection hyperparameters for \gls{TED} dev with 4-gram LM.}
	\label{fig:abl}	
	\vspace{-1.0em}
\includegraphics{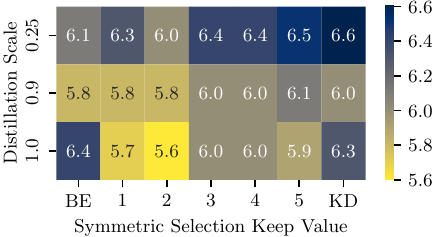}
\vspace{-2em}
\end{figure}
As a first study we present the results of symmetric selection evaluated on the \gls{TED} dev set over the three distillation scales 0.25, 0.9 and 1.0, and the symmetric selection parameter range of 1 to 5.
As visible in \Cref{fig:abl}, for a distillation scale of 0.25 there is a marginal improvement for the case of $n = 2$, but otherwise the performance degrades up to the baseline \gls{KD} performance. 
For a distillation scale of 0.9 the performance stagnates for lower keep values and reaches the baseline \gls{KD} performance at $n = 3$.
Interesting is the case of a scale of 1.0, where both the blank elimination and the baseline knowledge distillation perform worse than the best value of 5.6 \% for the case of $n = 2$.
This means that the model is able to benefit from the additional but limited amount of blanks included during distillation, producing a result that outperforms the other scales.
Not only does the introduction of symmetric blank selection improve the \gls{WER}, but additionally the dependence on any labels is dropped, as the model is only optimized on the teacher outputs. 

When analyzing the distributions, we find that on average 54\% of a sequence is determined to be non-blank by the teacher. 
Applying symmetric selection these values increase to 66\%, 69\%, 72\%, 74\% and 76\% ($1 \leq n \leq 5$). 
This means that in the largest case a quarter of the distilled positions is discarded, while overall three quarters of the sequence are considered for distillation.
Consequently, a quarter of the sequence consists of larger areas of blank positions, and they were not selected.

We combine these results with the other previously introduced approaches and extend the analysis to \gls{LBS} in \Cref{tab:blank2}.
For all cases where the modification of \gls{KD} has a hyperparameter, we present the best value for the corresponding set in the table, even though in all cases the ranges as stated in \Cref{sec:blank_sel} where tested.

Symmetric selection on \gls{LBS} is not able to outperform the simple \gls{KD} baseline, but overall stabilizes blank elimination for the distillation scales 0.9 and 1.0. 
The \gls{WER} of a distillation scale of 1.0 is even slightly better than for \gls{KD}, meaning that without labels keeping some blanks in a symmetric fashion is helpful. 
Trimming blanks before and after the sequence does not help for \gls{TED}, but produces a marginally better result than the \gls{KD} baseline for \gls{LBS}.
Selecting blanks based on a threshold or in a random fashion does not improve the distillation for \gls{TED}.
For \gls{LBS} we see a similar trend, with the exception of a distillation scale of 0.25. 
Here the models perform similar to the \gls{KD} baseline. 
From this we conclude that for \gls{LBS} and a distillation scale of 0.25 the position of the blank information is not important for the distillation, but rather a certain amount of blanks need to be seen during training for a good performance.

Overall we can conclude that when enabling symmetric selection during \gls{KD}, the dependence on the labeled data can be dropped and the model can be trained only on the teachers output, as visible in the results with a distillation scale of 1.0.
For \gls{TED} this symmetric selection without \gls{CTC} loss produced the best results, while for \gls{LBS} it helped reduce the gap to the results of training with a distillation scale of 0.25.

\begin{table}[t]
	\caption{Performance of blank selection mechanisms \gls{TED} and \gls{LBS} with 4-gram LM. For \gls{LBS} we evaluate on the \textit{other} portions of the evaluation data. In the first two rows $\dagger$ represents the distillation scale for \gls{TED} and $*$ the scale for \gls{LBS}.}
	\label{tab:blank2}
	\vspace{-5ex}
	\begin{center}
		\resizebox{3.2in}{!}{
			\begin{tabular}{|c|c||c|c|c||c|c|c|}
				\hline
				\multirow{3}{*}{\makecell{Blank \\ Select.}} & \multirow{3}{*}{\makecell{KD \\ Scale}} & \multirow{3}{*}{\makecell{Hyper- \\ param.}} & \multicolumn{2}{c||}{\gls{TED}} &\multirow{3}{*}{\makecell{Hyper- \\ param.}} &\multicolumn{2}{c|}{\gls{LBS}} \\ 
				\cline{4-5}
				\cline{7-8}
				& & &  \multicolumn{2}{c||}{WER [\%]} & & \multicolumn{2}{c|}{WER [\%]}\\
				\cline{4-5}
				\cline{7-8}
				& & & dev & test & & dev & test \\
				\hline
				- & \multirow{2}{*}{\makecell{0.9$^\dagger$ \\/ 0.25$^*$}} &\multirow{2}{*}{-} & 6.0 & 6.7 &\multirow{2}{*}{-} & \textbf{5.6} & 6.1\\
				\cline{1-1}
				\cline{4-5}
				\cline{7-8}
				\multirow{1}{*}{Bl. Elim.} & & & 5.8 & 6.5 & & 5.7 & 6.1\\
				\hline
				\hline
				\multirow{3}{*}{\makecell{Sym.}}& 0.25 & $n = 2$ & 6.0 & 6.8 & $n = 4$ & 5.7 & 6.0\\
				\cline{2-8}
				& 0.9 & $n = 1$ & 5.8 & \textbf{6.3} & \multirow{2}{*}{$n = 3$} & 5.8 & 6.2\\
				\cline{2-5}
				\cline{7-8}
				& 1.0 & $n = 2$& \textbf{5.6} & \textbf{6.3} & & 5.8 & 6.2 \\
				\hline
				\multirow{3}{*}{\makecell{Trim}}& 0.25 & \multirow{3}{*}{-} & 6.5 & 6.8 &\multirow{3}{*}{-} & \textbf{5.6} & \textbf{5.9} \\
				\cline{2-2}
				\cline{4-5}
				\cline{7-8}
				& 0.9 & & 6.1 & 6.5& & 5.9 & 6.3 \\
				\cline{2-2}
				\cline{4-5}
				\cline{7-8}
				& 1.0 & & 6.2 & 6.6& & 6.0 & 6.4 \\
				\hline				
				\multirow{3}{*}{Thresh.}&  0.25& $\alpha = 0.9$ & 6.1 & 7.0 & $\alpha = 0.8$ & \textbf{5.6} & 6.1\\
				\cline{2-8}
				& 0.9 & $\alpha = 0.8$ & 5.9 & 6.4 & \multirow{2}{*}{$\alpha = 0.9$} & 5.9 & 6.4 \\
				\cline{2-5}
				\cline{7-8}
				& 1.0 & $\alpha = 0.9$ & 6.0 & 6.5 & & 6.2 & 6.6 \\
				\cline{2-8}
				\hline
				\multirow{3}{*}{Random}&  0.25& \multirow{2}{*}{$\beta = 1.0$} & 6.3 & 7.0 & \multirow{2}{*}{$\beta = 0.5$} & \textbf{5.6} & 6.1 \\
				\cline{2-2}
				\cline{4-5}
				\cline{7-8}
				& 0.9 & & 6.0 & 6.6 & & 6.0 & 6.2 \\
				\cline{2-8}
				& 1.0 & $\beta = 0.5$ & 6.1 & 6.4 & $\beta = 2.0$ & 6.0 & 6.4 \\
				\cline{2-8}
				\hline
			\end{tabular}
		}
	\end{center}
	\vspace{-6ex}
\end{table}

\vspace{-0.75em}
\section{Limitation and Future Work}
One of the caveats of most of our blank aware approaches to improving \gls{CTC}-based \gls{KD} is the introduction of an additional hyperparameter to tune, as our experiments show different settings are optimal for different corpora and distillation scales.
This might be due to the fact that blank distributions induced by \gls{CTC} training might depend a lot on the data, as blank serves both as a wait and a silence token. 
Consequently, corpora with different amounts of silence will have largely different blank distributions.
Future work on this topic should aim to remove additional hyperparameters by introducing a mechanism which decides the blanks to be kept in a more automatic fashion.
Thresholding is a step in that direction, but still has a different (although easier to tune) parameter.
Furthermore, it should be investigated how to deal with the different blank amounts by potentially taking the induced blank distribution of the corpus into account before blank selection.
This way different corpora with different blank amounts could behave in a more similar fashion reducing tuning efforts.

\vspace{-0.75em}
\section{Conclusion}
In this work we analyzed the role of blank symbols during knowledge distillation for \gls{CTC}-based \gls{ASR}. 
For this we used a large pre-trained teacher to distill knowledge to a medium sized student for two different English corpora. 
We show that baseline extensions as blank elimination do not always improve the model performance and a dependence on the target labels is important to produce good results. 
We analyzed different new ways of selecting blanks for distillation, namely trimming, thresholding and random selection.
These methods showed no improvements over the baseline on LibriSpeech, while degrading on TED-LIUMv2.
Lastly, we introduced symmetric selection, selecting blank positions close to the non blank positions during distillation, while ignoring further away blanks. 
With a distillation scale of 1.0 and symmetric blank selection, we outperform the blank elimination method on TED-LIUMv2 and close the gap to knowledge distillation with a distillation scale of 0.25 on LibriSpeech.
Through this, symmetric selection enables training on unsupervised data, as a CTC loss is no longer required for good performance.

\section{Acknowledgments}
This work was partially supported by NeuroSys, which as part of the initiative “Clusters4Future” is funded by the Federal Ministry of Education and Research BMBF (funding IDs 03ZU2106DA and 03ZU2106DD), and by the project RESCALE within the program \textit{AI Lighthouse Projects for the Environment, Climate, Nature and Resources} funded by the Federal Ministry for the Environment, Nature Conservation, Nuclear Safety and Consumer Protection (BMUV), funding ID: 67KI32006A.

\bibliographystyle{IEEEtran}
\bibliography{mybib}

\end{document}